\pdfoutput=1

\documentclass[11pt]{article}

\usepackage[]{EMNLP2022}

\usepackage{times}
\usepackage{latexsym}
\usepackage[T1]{fontenc}

\usepackage[utf8]{inputenc}

\usepackage{microtype}

\usepackage{inconsolata}
\usepackage{times}
\usepackage{latexsym}
\usepackage{algorithm}
\usepackage{algorithmic}
\usepackage[switch]{lineno} 
\usepackage{CJKutf8}
\usepackage{multirow}
\usepackage{booktabs}
\usepackage{graphicx}
\usepackage{amsmath}
\usepackage{amssymb}
\usepackage{xpinyin}
\usepackage{pifont}
\usepackage{subfigure}
\usepackage{xcolor}
\usepackage{longtable}
\definecolor{occasion}{RGB}{255, 108, 13}
\definecolor{object}{RGB}{0, 176, 240}
\newcommand{\DatasetName}{EBleT}

\title{Towards Attribute-Entangled Controllable Text Generation: \\ A Pilot Study of Blessing Generation}

\author{
Shulin Huang$^{1}$\thanks{$^*$ indicates equal contribution.},~Shirong Ma$^{1*}$,~Yinghui Li$^{1}$,~Yangning Li$^{1}$,\\~\textbf{~Shiyang Lin}$^{2}$,~\textbf{Hai-Tao Zheng}$^{1,3}$\thanks{ $^{\dagger}$ Corresponding author: Hai-Tao Zheng and Ying Shen. (E-mail: zheng.haitao@sz.tsinghua.edu.cn, sheny76@mail.sysu.edu.cn)} \and \textbf{Ying Shen}$^{2\dagger}$\\
        $^{1}$Tsinghua Shenzhen International Graduate School, Tsinghua University \\ 
        $^{2}$School of Intelligent Systems Engineering, Sun-Yat Sen University\\
        $^{3}$Peng Cheng Laboratory \\
        \texttt{\{sl-huang21, masr21\}@mails.tsinghua.edu.cn}
}

\begin{document}
\maketitle
\begin{abstract}
Controllable Text Generation (CTG) has obtained great success due to its fine-grained generation ability obtained by focusing on multiple attributes.
However, most existing CTG researches overlook \emph{how to utilize the attribute entanglement to enhance the diversity of the controlled generated texts}. 
Facing this dilemma, we focus on a novel CTG scenario, i.e., \textbf{blessing generation} which is challenging because high-quality blessing texts require CTG models to comprehensively consider the entanglement between multiple attributes (e.g., objects and occasions). 
To promote the research on blessing generation, we present \DatasetName{}, a large-scale \textbf{E}ntangled \textbf{Ble}ssing \textbf{T}ext dataset containing 293K English sentences annotated with multiple attributes. 
Furthermore, we propose novel evaluation metrics to measure the quality of the blessing texts generated by the baseline models we designed.
Our study opens a new research direction for controllable text generation and enables the development of attribute-entangled CTG models. Our dataset and source codes are available at \url{https://github.com/huangshulin123/Blessing-Generation}.
\end{abstract}

\section{Introduction}
Controllable Text Generation (CTG) aims to automatically generate the text under the restrictions of given conditions~\cite{prabhumoye-etal-2020-exploring,DBLP:journals/corr/abs-2112-11739,sun2022non}. 
As the mainstream, controlling multiple attributes enriches the information contained by generation and matches the demand of application scenarios, such as generating Chinese poetry~\cite{yi2020mixpoet}, restaurant reviews~\cite{chen2021aspect}, and product descriptions~\cite{8946478}. 

\begin{figure}
    \centering
    \includegraphics[width=0.98\columnwidth]{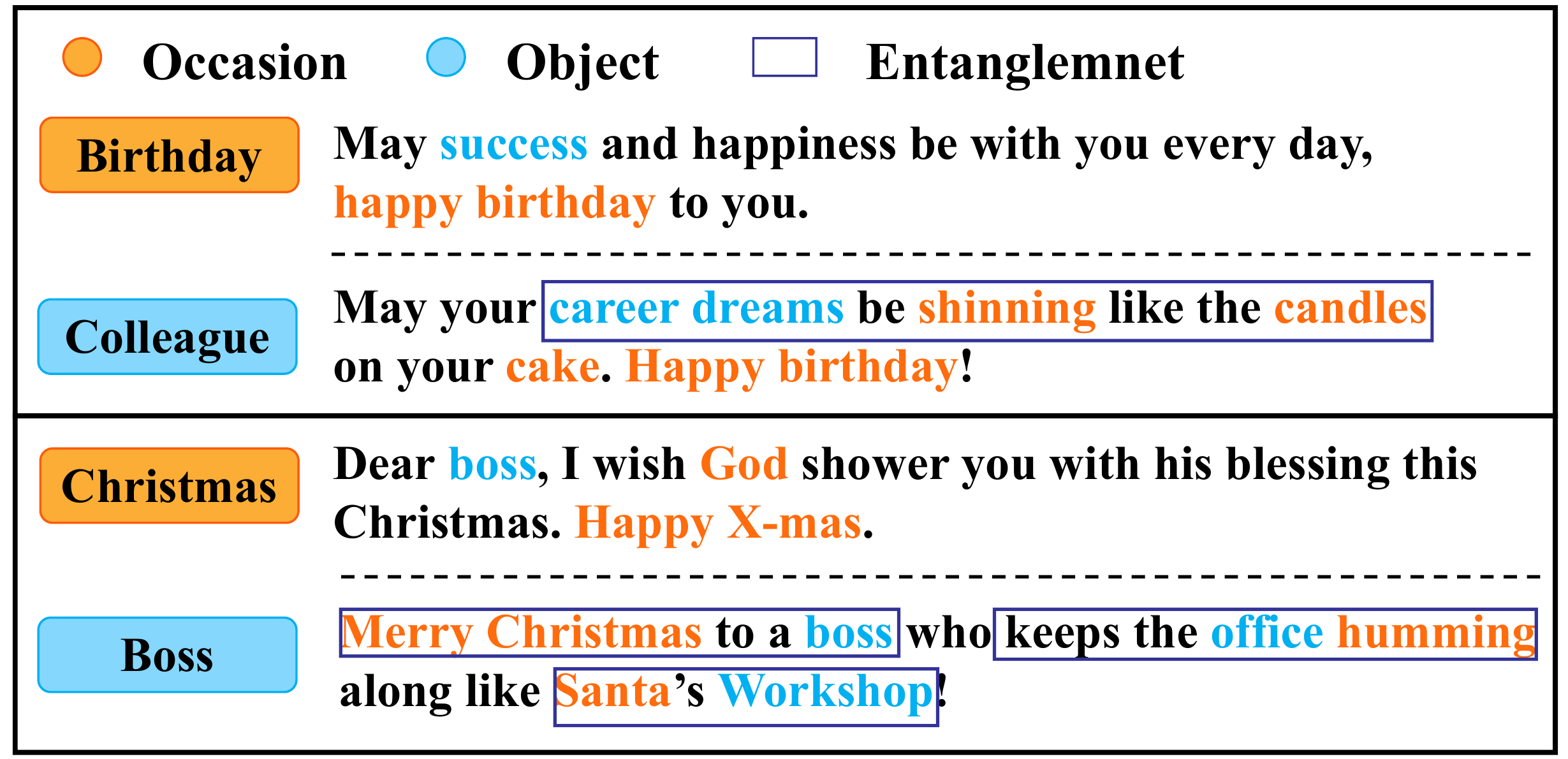}
    \caption{
    Two groups of blessing examples. Each group contains blessing messages without (top) and with (bottom) the attribute entanglement. \textbf{Representative elements} of \textcolor{occasion}{occasion}/\textcolor{object}{object} attributes are marked.
    }

    \label{fig:my_label}
\end{figure}

\begin{CJK*}{UTF8}{gbsn}
Take the Chinese poetry generation task as an example, one beautiful poetry sentence should contain multiple attributes and reflect the entanglement (or mixture) of them through reasonable connection, e.g., in the sentence “胡马南来路已荒 (The enemy's warhorses march to the south, through destroyed roads)”, “胡马 (enemy's warhorses)” is a representative element of the military career, “南来 (march to the south)” and “荒 (destroyed)” represent the attribute of troubled times. 
This poetry sentence vividly depicts a picture of War in troubled times through the entanglement of attributes in just seven characters.
\end{CJK*}
\citet{yi2020mixpoet} also claim that considering the entanglement among attributes can effectively enhance the quality and diversity of generated poetry. 
Therefore, we believe that \emph{better CTG models must focus on the effect of attribute entanglement, i.e.,
enhancing the reflection of multiple attributes through the use of various representative elements in the generated text.}

For Chinese CTG, with poetry generation as a typical scenario, researchers have conducted in-depth research on attribute entanglement, but in the English CTG field, the research on attribute entanglement has not been explored.
Therefore, to promote research on attribute-entangled CTG in the English community, in this paper we focus on \textbf{blessing generation}, a new CTG task that plays a key role in social scenarios. 
The automatically generated blessings will greatly promote interpersonal communication and enrich people's daily life.
More crucially, the blessing generation task is challenging due to its high requirement for entanglement between attributes, such as objects and occasions. 
As shown in Figure~\ref{fig:my_label}, 
“Santa's Workshop” connects the occasion (Christmas) and object (Boss) into one phrase, making the blessing wonderful. 
A more vivid blessing embodies these two attributes in an intertwined manner, such as “keeps the office humming along like Santa's Workshop”.

Facing the vacant of blessing generation, we construct \DatasetName{}, a large-scale \textbf{E}ntangled \textbf{Ble}ssing \textbf{T}ext dataset annotated with multiple attributes. 
Particularly, the \DatasetName{} is constructed with the following two features: (1) \DatasetName{} contains 23 occasions and 34 objects annotated on 293,403 blessing texts from 12 blessing websites. (2) As 92\% of the blessing texts are personalized for the corresponding attributes, \DatasetName{} has at least 82\% data containing the entanglement between attributes.

Additionally, the common generation evaluation metrics cannot reflect the characteristics of blessings clearly. To evaluate the generated blessings more comprehensively, we propose novel metrics to automatically calculate the degree of attribute entanglement and the quality of blessings. 
Our experiments demonstrate that mainstream CTG methods struggle to contain the entanglement. 
Moreover, existing methods can not balance the fluency, diversity, and entanglement between attributes. 
These results indicate that the blessing generation task we focus on is challenging and could serve as a useful benchmark for CTG research.

\section{Task Definition} 
The blessing generation task aims to obtain a generation model $\mathcal{G}(x_1,x_2;\theta)$ parameterized by $\theta$.
Given the input attributes containing an object $x_1\in X_1$ and an occasion $x_2\in X_2$, the model $\mathcal{G}$ should output a blessing text $y$ sent to $x_1$ for $x_2$, where $y=\{y_1, y_2,..., y_n\}$ is a sequence containing $n$ words, and $x_i(i=1,2)$ is a word or a phrase belonging to a collection of objects or occasions. The generated text $y$ should reflect not only the language style of blessing, but also effective entanglement between both attributes.
Additionally, the evaluation metrics for the language style of blessing and entanglement are described in Section~\ref{sec:eval}.

\section{\DatasetName{} Dataset}

\subsection{Dataset Construction}

\paragraph{Data Collection} 
We search blessing-related keywords (e.g., “send blessing”, “send wish”) via Google Search and obtain 12 blessing websites.
We check the licences of those websites to ensure that data from these websites can be legally employed for our non-profit academic research.
The occasions and objects are labeled by page headings and subheadings from these websites. 
Therefore, we obtain the headings and subheadings, as well as corresponding lists of blessing texts. The occasions and objects are extracted from the headings and subheadings.
We totally collect about 1 million texts from the web as the raw corpus.

\paragraph{Data Cleaning} 
After acquiring the original corpus, we remove completely duplicate sentences, delete all non-English text, and remove the sentences that do not reflect corresponding occasion/object attributes.
Additionally, we observe that too long or too short sentences are mostly noise. Therefore, to further clean the dataset, we keep only sentences in the range of 10 to 200 words in length.

\paragraph{Human Evaluation}
To manually evaluate the quality of \DatasetName{}, we randomly select 20 data samples from each “object-occasion” pair except for the pairs related to the “General” object and finally obtain 5,520 data samples.
Then we employ 3 college students who are English native speakers as annotators to manually assess the personalization and entanglement scores of these samples. As the annotation payment, we provide them 5 dollars for every 100 sentences they judged.
Besides, to ensure the reliability of their scores, we carefully explain the concept of personalization and entanglement to them before the start of annotation.
Specifically, \textbf{a blessing can be called personalized if the annotator can easily know its labeled occasion/object. Moreover, a blessing can be called entangled if it cleverly blends the characteristics of the labeled occasion/object, rather than combining the two so rigidly that it can be substituted for any other occasions or objects.}
After being familiar with the concepts of personalization and entanglement, our annotators are asked to judge the sampled data and give the score (0 - common, 1 - personalized, 2 - both personalized and entangled). 
We take the majority vote as the annotation result for a data sample.
The Fleiss’ kappa~\cite{fleiss1971measuring} of the annotations is 0.837, which indicates the annotation results of our annotators can be regarded as “almost perfect agreement”~\cite{landis1977measurement}. The results of human evaluation will be presented and analyzed in the “Dataset Quality” of Section~\ref{Dataset_Analysis}. 

\subsection{Dataset Analysis}
\label{Dataset_Analysis}
\paragraph{Dataset Statistics} Table~\ref{tab:data_statistics} describes statistics of \DatasetName{}.
Compared with previous annotated CTG datasets, e.g., ROCStories~\cite{mostafazadeh-etal-2016-corpus} with 50K stories, GYAFC~\cite{rao-tetreault-2018-dear} with 53K sentences and ToTTo~\cite{parikh-etal-2020-totto} with 121K tables, our \DatasetName{} containing 293K blessing texts with corresponding occasion and object labels can be regarded as a sufficiently large-scale dataset. Moreover, our dataset consists of up to 276 pairs crossed by 23 categories of occasions and 34 categories of objects, which is challenging for models to learn the characteristics of each category of occasions and objects and to entangle them.
More details and examples of \DatasetName{} are shown in Appendix~\ref{sec:datadetail}.
\begin{table}[ht]
\small
\centering
\begin{tabular}{lc}
\toprule
Property & Value \\ \midrule
Dataset Size & 293,403 \\
Average Length &  43.06 \\
\# Occasions & 23 \\
\# Objects   & 34 \\
\# Occasion-object Pairs & 276 \\
\bottomrule
\end{tabular}
\caption{Dataset statistics of \DatasetName{}.}
\label{tab:data_statistics}
\end{table}

\paragraph{Dataset Quality}
Table~\ref{tab:human_evaluation} shows human evaluation results of \DatasetName{}. It indicates that about 92\% of the blessing texts are personalized for the corresponding attributes, and about 82\% data reflect the entanglement between attributes, which demonstrates the quality of \DatasetName{}.

\begin{table}[ht]
\small
\centering
\begin{tabular}{ll|ccc}
\toprule
\multicolumn{1}{c}{} & & \#Sample & \#Per. & \#Ent. \\ \midrule
\multirow{2}{*}{Occasion} & Christmas & 480 & 445 & 397 \\
 & Halloween & 160 & 144 & 134 \\ \midrule
\multirow{2}{*}{Object} & Teacher & 200 & 181 & 164 \\
 & Boss & 140 & 126 & 118 \\ \midrule
 & Total & 5,060 & 4,676 & 4,170 \\
\bottomrule
\end{tabular}
\caption{Partial human evaluation results of~\DatasetName{}. \#Sample, \#Per. and \#Ent. denote the total number of sampled sentences, the number of personalized sentences and the number of entangled sentences respectively. The full list is presented in Table~\ref{tab:human2}.}
\label{tab:human_evaluation}
\end{table}

\begin{figure}[h]
    \centering
    \includegraphics[width=0.80\columnwidth]{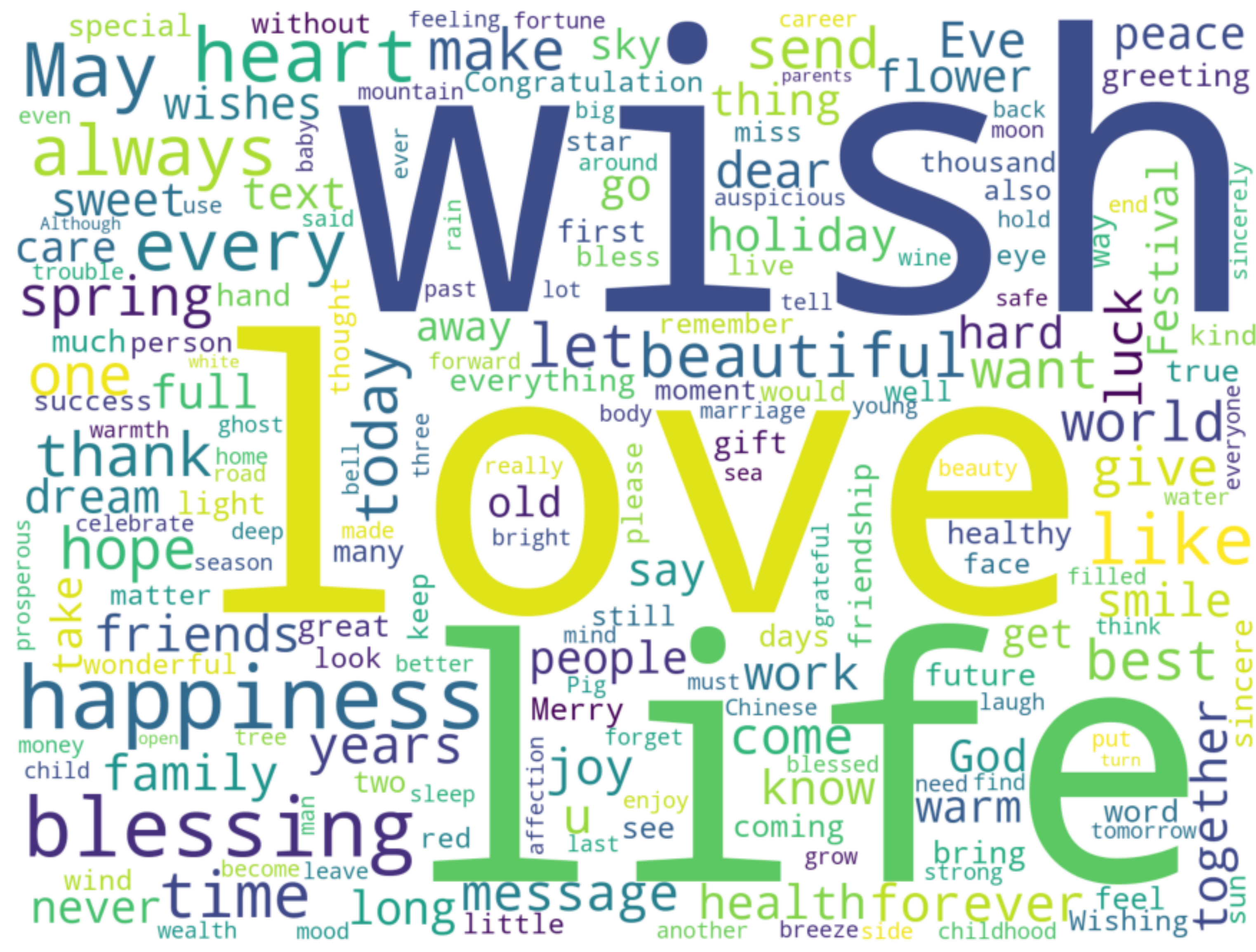}
    \caption{The word cloud visualization of \DatasetName{}.}
    \label{fig:word_cloud}
\end{figure}

\paragraph{Dataset Visualization} 
After removing the stopwords and the words related to specific occasions and objects, we plot the word cloud of \DatasetName{} as shown in Figure~\ref{fig:word_cloud}.
We find out that some words (e.g., “wish”, “love”, and “happiness”) appear frequently. 
This phenomenon not only meets our common sense, i.e., blessing texts usually express wishes for each other, but also provides a class of words that need to be focused on for the development of future blessing generation models.

\section{Evaluation Metrics}\label{sec:eval}
\subsection{Blessing Score}
To measure the quality of blessings, Blessing Score should reflect the extent to which a sentence fits the language style of the blessing.
By counting word frequency, we observe that some words, e.g., “happy”, “merry”, and “heart”, frequently appear in blessing texts rather than in other texts.
We obtain the 50 most frequently occurring words and remove the stopwords. These words are utilized to construct the bag-of-words of the blessing $B$.

For a sentence to be evaluated, to avoid the influence of irrelevant words, we use KeyBERT~\cite{grootendorst2020keybert} to extract 10 keywords to form a keyword list $K$ as a representative of the sentence. All words in $B$ and $K$ are converted to word embeddings by Word2Vec~\cite{DBLP:conf/nips/MikolovSCCD13} model $E(.)$. For each keyword, we calculate its maximum similarity to all words in $B$, and then average the maximum similarity of all keywords to obtain the Blessing Score (BLE). It is formulated as follows:

{
\small
\begin{equation}\label{eq:bs}
    \text{BLE}=\frac{1}{|K|}\sum_{\substack{w \in K}}\max_{b \in B}{\frac{E(w)\cdot E(b)}{\Vert E(w) \Vert\cdot \Vert E(b)\Vert}}.
\end{equation}
}

\subsection{Entanglement Score}
To evaluate the degree of attribute entanglement, we assume that a blessing sentence with higher Entanglement Score should satisfy that the elements related to the occasions and objects appear simultaneously in more clauses. Further, occasion-related and object-related elements should alternate more times in one more entangled blessing sentence.

We construct two bags-of-words $B_1, B_2$ to represent the occasion-related and object-related elements respectively. Specifically, the bags-of-words contain words directly related to the corresponding occasions and objects, which are listed in Table~\ref{tab:bag} and Table~\ref{tab:bag2} of the Appendix.

For the Entanglement Score, we calculate whether words related to the two attributes occur simultaneously within each clause by cosine similarity, and add a bonus term $O$\footnote{The specific implementation of our designed bonus is presented in the source code of the supplementary material.} for the cases where related words occur alternately multiple times. Formally, for each sentence $S$ to be evaluated, we split $S$ into $m$ clauses $S=\{s_1,s_2,...,s_m\}$ and each clause $s_i$ consists of $n$ words $s_i=\{w_{i1},w_{i2},...,w_{in}\}$.
The Entanglement Score (ENT) for $S$ is calculated as follows:

{
\small
\begin{gather}
    \text{ENT}=\sum_{\substack{s_i \in S}} I((\exists\ w_{ij},w_{ik} \in s_i)\ C(w_{ij},w_{ik}))+O,\\
    C(w_1,w_2)=\text{sim}(w_1,B_1)>t\ \land \ \text{sim}(w_2,B_2)>t, \\
    \text{sim}(w,B)=\max_{\substack{b \in B}}\frac{E(w)\cdot E(b)}{\Vert E(w)\Vert \cdot \Vert E(b)\Vert},
\end{gather}
}
where $I(c)$ is the indicator function, which has a value of 1 when the condition $c$ is satisfied, $t$ is a predetermined threshold.

\begin{figure}
    \centering
    \includegraphics[width=0.80\columnwidth]{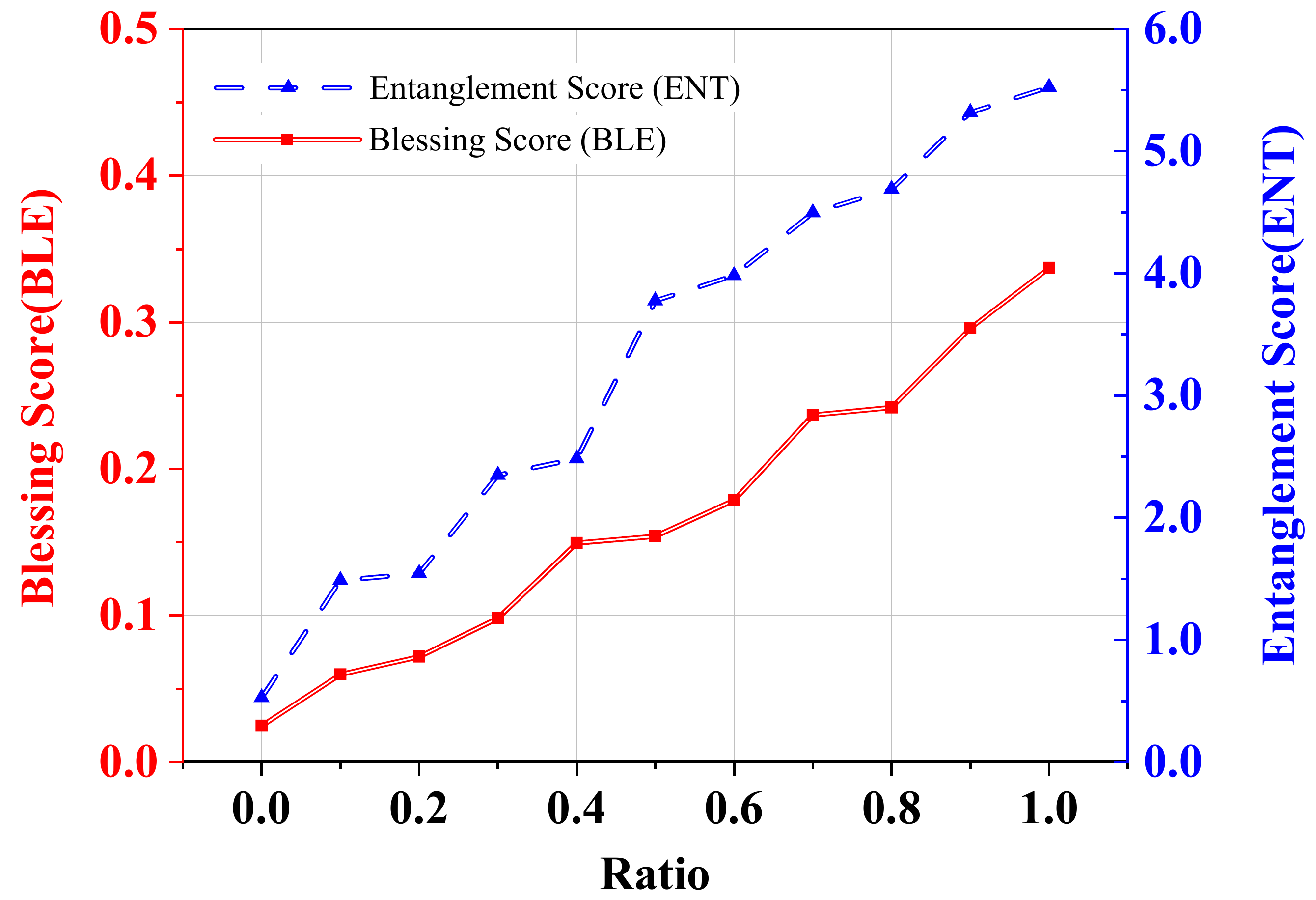}
    \caption{The correlation between human annotations and automatic metrics. The horizontal axis represents the proportion of the set that is manually annotated as blessing or entanglement. 
    }

    \label{fig:metric_verify}
\end{figure}

\begin{table*}[ht]
\small
\centering
\begin{tabular}{l|ccc|cccc|cc}
\toprule
 & BLEU$\uparrow$ & ROUGE-L$\uparrow$ & WMD$\downarrow$ & PPL$\downarrow$ & DIST-1$\uparrow$ & DIST-2$\uparrow$ & DIST-3$\uparrow$ & BLE$\uparrow$ & ENT$\uparrow$ \\
\midrule
Common & - & - & - & - & 0.609 & 0.953 & 0.998 & 0.022 & - \\
\midrule
GPT-2 & 0.225 & 0.382 & 1.018 & 20.12 & 0.176 & 0.327 & 0.405 & 0.308 & 3.52 \\
T5 & \textbf{0.247} & \textbf{0.393} & \textbf{1.015} & \textbf{13.47} & 0.140 & 0.274 & 0.358 & \textbf{0.334} & 3.23 \\
\midrule
GPT-2 + CVAE & 0.137 & 0.340 & 1.058 & 28.84 & \textbf{0.409} & \textbf{0.788} & \textbf{0.907} & 0.223 & 3.63 \\
GPT-2 + Adv. & 0.147 & 0.349 & 1.050 & 28.32 & 0.397 & 0.778 & 0.903 & 0.226 & \textbf{3.78} \\
\midrule
Reference & - & - & - & - & 0.455 & 0.830 & 0.928 & - & 4.54 \\
\bottomrule
\end{tabular}
\caption{Performance of different models on \DatasetName{}. \textbf{Common} represents the news texts collected from British Broadcasting Corporation which is used to make the comparison with blessings. "$\uparrow$" represents higher is better for this metric and "$\downarrow$" represents lower is better.}
\label{tab:res}
\end{table*}

\subsection{Metric Verification}
To verify the effectiveness of our proposed blessing and entanglement score, we conduct consistency analyses between automatic scores and human annotations.
We extract 11 subsets and each of them has 100 pieces of data. 
Meanwhile, we make the proportion of blessings or entanglement (annotated by humans) in each set different, which is from 0.0 to 1.0.
The average blessing score and entanglement score for the 11 subsets are calculated by our metrics.
The results presented in Figure~\ref{fig:metric_verify} demonstrate that our proposed metrics are highly consistent with the results of manual annotation.

\section{Experiments}

\subsection{Experiment Setup}
We set up experiments to evaluate the performance of existing models to generate entangled blessing texts. The full dataset is divided into a training set, a validation set and a test set in the ratio of 9:0.5:0.5 by stratified sampling.

To measure the consistency of generated outputs and reference blessing texts, we utilize BLEU~\cite{papineni2002bleu} and WMD~\cite{kusner2015word}. 
WMD is a method to calculate the minimum embedded word distance required for a document to transfer to another one. 
In addition, we use Perplexity and Distinct-n(n=1,2,3)~\cite{li2016diversity} to evaluate the fluency and diversity of generated outputs. Specifically, GPT-Neo~\cite{gao2020pile} is employed as the language model to obtain the perplexity.
Furthermore, we use Blessing Score and Entanglement Score mentioned in Section~\ref{sec:eval} to evaluate the quality of blessings.

We evaluate two widely used generation models on \DatasetName{} for our proposed task: 

\textbf{GPT-2}~\cite{radford2019language} is a Transformer-based decoder-only model~\cite{liu2022we} which achieves stable and excellent generation performance. 
For this task, we design a prompt: “Send this blessing to \textit{<object>} for \textit{<occasion>}”, where \textit{<object>} and \textit{<occasion>} represent the object and occasion attributes, respectively. The prompt is utilized for the prefix input of GPT-2 model.
Diverse Beam Search~\cite{DBLP:journals/corr/VijayakumarCSSL16} is employed as the decoding method during the generation process to ensure diversity of generated blessings.

\textbf{T5}~\cite{raffel2020exploring} is a model of the encoder-decoder framework which is commonly used for text-to-text generation tasks. 
The prompt mentioned above is utilized for the input of encoder side of T5 model.

Additionally, we consider applying CVAE~\cite{DBLP:conf/nips/SohnLY15} for generation and using the latent variables to represent the entanglement of the two input attributes. Following the previous work~\cite{fang2021transformer}, we employ pretrained GPT-2 as the backbone of CVAE to obtain higher quality generated results. Furthermore, we employ adversarial training (Adv.)~\cite{yi2020mixpoet} instead of minimizing KL divergence in CVAE to allow the model to learn more complex entangled representations.

\subsection{Experiment Results}

The results of Table~\ref{tab:res} demonstrate that:
(1) Models trained on \DatasetName{} can generate fluent blessing texts. The language style of generated texts is generally consistent with that of the blessing texts in the dataset.
(2) The diversity and Entanglement Score of texts generated by GPT-2 and T5 are actually low. Meanwhile, employing CVAE or adversarial training architecture based on GPT-2 can effectively improve these two metrics but slightly reduce the quality of blessing.
Additionally, the architecture of adversarial training outperforms CVAE in the entanglement and the quality of blessing, suggesting that the adversarial training architecture is more appropriate for entangling the attributes into generation.
(3) There exists a gap of diversity and Entanglement Score between generated texts and references.
It indicates that \DatasetName{} is a challenging benchmark for exploring the entanglement of attributes in CTG. Future work on this task should consider all the metrics of fluency, diversity, quality of blessings, and entanglement to generate blessings that are more in line with human expression.

\section{Related Work}
Controllable text generation (CTG) usually takes the controlled element and source text (which can be missing) as the input. Based on the input, the generation model produces the target text satisfying controlled elements. 
According to the core of CTG, i.e., the diversified controlled elements, we can divide CTG into the following two categories:

\textbf{Attribute Control}: \citet{ghosh2017affect} add the sentiment information into the generator to control the sentiment of the generated sentences. \citet{luo2019learning} explore a framework including sentiment analysis and sentiment generator to control the fine-grained sentiment of generation. \citet{chen2021aspect} introduce a mutual learning framework to generate emotionally controllable comments. In addition, \citet{wang2019harnessing} control the style of the generated text to present a specific style of writing. \citet{zhang2018personalizing} build a generation system to generate conversations with the specific persona.

\textbf{Content Control}: \citet{cao2015novel} control the topic of generation, exploring the latent semantics of vocabularies and texts to get the distribution of the topic. \citet{keskar2019ctrl} add different controlling code to realize topic control. \citet{koncel2016theme} use the generator to edit the articles written by humans, changing the theme without changing the original story. Additionally, \citet{zheng2020controllable} build LSTML and LSTMR to make sure the entities appear in the generated summary. \citet{xu2020megatron} incorporate keywords into each sentence of the story over the generation process. \citet{kikuchi2016controlling, duan2020pre} introduce the methods for controlling the output sequence length.

However, existing research work on controlled generation doesn't include the work related to blessing and neglects the entanglement among attributes. Blessings can be used in many aspects of life, such as e-cards, advertisements, and so on. Thus we introduce a new task - blessing generation and propose the corresponding dataset \DatasetName{}.

\section{Conclusion}
To explore the entanglement between attributes, we present \DatasetName{}, a blessing dataset that presents a new controllable generation task. We propose novel metrics to automatically measure attribute entanglement and the quality of blessings. We also provide several baselines and conduct experiments for blessing generation. Experimental results demonstrate that \DatasetName{} could serve as a useful benchmark for attribute entanglement in CTG.

\section*{Limitations}
In this paper, we conduct experiments on \DatasetName{} employing some representative mainstream models.
Since our work is only a pilot study of attributed-entangled CTG, we do not conduct experiments on more controllable generation models. 
Because of the challenge of \DatasetName{}, we suggest that more complex models can be implemented for improving the performance of blessing generation.

\section*{Ethical Considerations}
In this paper, to facilitate the study of attribute-entangled CTG, we propose the blessing generation task which needs to pay attention to the attribute entanglement to obtain vivid blessings. We believe that the blessing generation task embodies humanistic care, and the various generated blessing texts can not only enrich people's daily life, but also promote interpersonal relationships.
We also present \DatasetName{}, a large-scale annotated blessing dataset. All the corpora used in \DatasetName{} come from freely available resources on public websites and do not involve any sensitive or illegal data.
Additionally, we design new automatic evaluation metrics to measure the quality of blessings. We think that our designed metrics are instructive for future research on the CTG tasks. After all, in the current CTG field, how to conduct an effective evaluation is also an important and yet unsolved problem.

\section*{Acknowledgement}
This research is supported by National Natural Science Foundation of China (Grant No.62276154 and 62011540405), Beijing Academy of Artificial Intelligence (BAAI), the Natural Science Foundation of Guangdong Province (Grant No. 2021A1515012640), Basic Research Fund of Shenzhen City (Grant No. JCYJ20210324120012033 and JSGG20210802154402007), and Overseas Cooperation Research Fund of Tsinghua Shenzhen International Graduate School  (Grant No. HW2021008).

\bibliography{anthology,custom}

\begin{thebibliography}{34}
\expandafter\ifx\csname natexlab\endcsname\relax\def\natexlab#1{#1}\fi

\bibitem[{Cao et~al.(2015)Cao, Li, Liu, Li, and Ji}]{cao2015novel}
Ziqiang Cao, Sujian Li, Yang Liu, Wenjie Li, and Heng Ji. 2015.
\newblock A novel neural topic model and its supervised extension.
\newblock In \emph{Proceedings of the AAAI Conference on Artificial
  Intelligence}, volume~29.

\bibitem[{Chen et~al.(2021)Chen, Lin, Qi, Hu, Li, Zhou, and
  Sun}]{chen2021aspect}
Huimin Chen, Yankai Lin, Fanchao Qi, Jinyi Hu, Peng Li, Jie Zhou, and Maosong
  Sun. 2021.
\newblock Aspect-level sentiment-controllable review generation with mutual
  learning framework.
\newblock In \emph{Proceedings of the AAAI Conference on Artificial
  Intelligence}, volume~35, pages 12639--12647.

\bibitem[{Dong et~al.(2021)Dong, Li, Gong, Chen, Li, Shen, and
  Yang}]{DBLP:journals/corr/abs-2112-11739}
Chenhe Dong, Yinghui Li, Haifan Gong, Miaoxin Chen, Junxin Li, Ying Shen, and
  Min Yang. 2021.
\newblock \href {http://arxiv.org/abs/2112.11739} {A survey of natural language
  generation}.
\newblock \emph{CoRR}, abs/2112.11739.

\bibitem[{Duan et~al.(2020)Duan, Xu, Pei, Han, and Li}]{duan2020pre}
Yuguang Duan, Canwen Xu, Jiaxin Pei, Jialong Han, and Chenliang Li. 2020.
\newblock Pre-train and plug-in: Flexible conditional text generation with
  variational auto-encoders.
\newblock In \emph{Proceedings of the 58th Annual Meeting of the Association
  for Computational Linguistics}, pages 253--262.

\bibitem[{Fang et~al.(2021)Fang, Zeng, Liu, Bo, Dong, and
  Chen}]{fang2021transformer}
Le~Fang, Tao Zeng, Chaochun Liu, Liefeng Bo, Wen Dong, and Changyou Chen. 2021.
\newblock Transformer-based conditional variational autoencoder for
  controllable story generation.
\newblock \emph{arXiv e-prints}, pages arXiv--2101.

\bibitem[{Fleiss(1971)}]{fleiss1971measuring}
Joseph~L Fleiss. 1971.
\newblock Measuring nominal scale agreement among many raters.
\newblock \emph{Psychological bulletin}, 76(5):378.

\bibitem[{Gao et~al.(2020)Gao, Biderman, Black, Golding, Hoppe, Foster, Phang,
  He, Thite, Nabeshima et~al.}]{gao2020pile}
Leo Gao, Stella Biderman, Sid Black, Laurence Golding, Travis Hoppe, Charles
  Foster, Jason Phang, Horace He, Anish Thite, Noa Nabeshima, et~al. 2020.
\newblock The pile: An 800gb dataset of diverse text for language modeling.
\newblock \emph{arXiv preprint arXiv:2101.00027}.

\bibitem[{Ghosh et~al.(2017)Ghosh, Chollet, Laksana, Morency, and
  Scherer}]{ghosh2017affect}
Sayan Ghosh, Mathieu Chollet, Eugene Laksana, Louis-Philippe Morency, and
  Stefan Scherer. 2017.
\newblock Affect-lm: A neural language model for customizable affective text
  generation.
\newblock In \emph{Proceedings of the 55th Annual Meeting of the Association
  for Computational Linguistics (Volume 1: Long Papers)}, pages 634--642.

\bibitem[{Grootendorst(2020)}]{grootendorst2020keybert}
Maarten Grootendorst. 2020.
\newblock \href {https://doi.org/10.5281/zenodo.4461265} {Keybert: Minimal
  keyword extraction with bert.}

\bibitem[{Keskar et~al.(2019)Keskar, McCann, Varshney, Xiong, and
  Socher}]{keskar2019ctrl}
Nitish~Shirish Keskar, Bryan McCann, Lav~R Varshney, Caiming Xiong, and Richard
  Socher. 2019.
\newblock Ctrl: A conditional transformer language model for controllable
  generation.

\bibitem[{Kikuchi et~al.(2016)Kikuchi, Neubig, Sasano, Takamura, and
  Okumura}]{kikuchi2016controlling}
Yuta Kikuchi, Graham Neubig, Ryohei Sasano, Hiroya Takamura, and Manabu
  Okumura. 2016.
\newblock Controlling output length in neural encoder-decoders.
\newblock In \emph{Proceedings of the 2016 Conference on Empirical Methods in
  Natural Language Processing}, pages 1328--1338.

\bibitem[{Koncel-Kedziorski et~al.(2016)Koncel-Kedziorski, Konstas,
  Zettlemoyer, and Hajishirzi}]{koncel2016theme}
Rik Koncel-Kedziorski, Ioannis Konstas, Luke Zettlemoyer, and Hannaneh
  Hajishirzi. 2016.
\newblock A theme-rewriting approach for generating algebra word problems.
\newblock In \emph{Proceedings of the 2016 Conference on Empirical Methods in
  Natural Language Processing}, pages 1617--1628.

\bibitem[{Kusner et~al.(2015)Kusner, Sun, Kolkin, and
  Weinberger}]{kusner2015word}
Matt Kusner, Yu~Sun, Nicholas Kolkin, and Kilian Weinberger. 2015.
\newblock From word embeddings to document distances.
\newblock In \emph{International conference on machine learning}, pages
  957--966. PMLR.

\bibitem[{Landis and Koch(1977)}]{landis1977measurement}
J~Richard Landis and Gary~G Koch. 1977.
\newblock The measurement of observer agreement for categorical data.
\newblock \emph{biometrics}, pages 159--174.

\bibitem[{Li et~al.(2016)Li, Galley, Brockett, Gao, and
  Dolan}]{li2016diversity}
Jiwei Li, Michel Galley, Chris Brockett, Jianfeng Gao, and William~B Dolan.
  2016.
\newblock A diversity-promoting objective function for neural conversation
  models.
\newblock In \emph{Proceedings of the 2016 Conference of the North American
  Chapter of the Association for Computational Linguistics: Human Language
  Technologies}, pages 110--119.

\bibitem[{Liu et~al.(2022)Liu, Li, Tao, Liang, and Zheng}]{liu2022we}
Ruiyang Liu, Yinghui Li, Linmi Tao, Dun Liang, and Hai-Tao Zheng. 2022.
\newblock Are we ready for a new paradigm shift? a survey on visual deep mlp.
\newblock \emph{Patterns}, 3(7):100520.

\bibitem[{Luo et~al.(2019)Luo, Dai, Yang, Liu, Chang, Sui, and
  Sun}]{luo2019learning}
Fuli Luo, Damai Dai, Pengcheng Yang, Tianyu Liu, Baobao Chang, Zhifang Sui, and
  Xu~Sun. 2019.
\newblock Learning to control the fine-grained sentiment for story ending
  generation.
\newblock In \emph{Proceedings of the 57th Annual Meeting of the Association
  for Computational Linguistics}, pages 6020--6026.

\bibitem[{Mikolov et~al.(2013)Mikolov, Sutskever, Chen, Corrado, and
  Dean}]{DBLP:conf/nips/MikolovSCCD13}
Tom{\'{a}}s Mikolov, Ilya Sutskever, Kai Chen, Gregory~S. Corrado, and Jeffrey
  Dean. 2013.
\newblock \href
  {https://proceedings.neurips.cc/paper/2013/hash/9aa42b31882ec039965f3c4923ce901b-Abstract.html}
  {Distributed representations of words and phrases and their
  compositionality}.
\newblock In \emph{Advances in Neural Information Processing Systems 26: 27th
  Annual Conference on Neural Information Processing Systems 2013. Proceedings
  of a meeting held December 5-8, 2013, Lake Tahoe, Nevada, United States},
  pages 3111--3119.

\bibitem[{Mostafazadeh et~al.(2016)Mostafazadeh, Chambers, He, Parikh, Batra,
  Vanderwende, Kohli, and Allen}]{mostafazadeh-etal-2016-corpus}
Nasrin Mostafazadeh, Nathanael Chambers, Xiaodong He, Devi Parikh, Dhruv Batra,
  Lucy Vanderwende, Pushmeet Kohli, and James Allen. 2016.
\newblock \href {https://doi.org/10.18653/v1/N16-1098} {A corpus and cloze
  evaluation for deeper understanding of commonsense stories}.
\newblock In \emph{Proceedings of the 2016 Conference of the North {A}merican
  Chapter of the Association for Computational Linguistics: Human Language
  Technologies}, pages 839--849, San Diego, California. Association for
  Computational Linguistics.

\bibitem[{Papineni et~al.(2002)Papineni, Roukos, Ward, and
  Zhu}]{papineni2002bleu}
Kishore Papineni, Salim Roukos, Todd Ward, and Wei-Jing Zhu. 2002.
\newblock Bleu: a method for automatic evaluation of machine translation.
\newblock In \emph{Proceedings of the 40th annual meeting of the Association
  for Computational Linguistics}, pages 311--318.

\bibitem[{Parikh et~al.(2020)Parikh, Wang, Gehrmann, Faruqui, Dhingra, Yang,
  and Das}]{parikh-etal-2020-totto}
Ankur Parikh, Xuezhi Wang, Sebastian Gehrmann, Manaal Faruqui, Bhuwan Dhingra,
  Diyi Yang, and Dipanjan Das. 2020.
\newblock \href {https://doi.org/10.18653/v1/2020.emnlp-main.89} {{ToTTo}: A
  controlled table-to-text generation dataset}.
\newblock In \emph{Proceedings of the 2020 Conference on Empirical Methods in
  Natural Language Processing (EMNLP)}, pages 1173--1186, Online. Association
  for Computational Linguistics.

\bibitem[{Prabhumoye et~al.(2020)Prabhumoye, Black, and
  Salakhutdinov}]{prabhumoye-etal-2020-exploring}
Shrimai Prabhumoye, Alan~W Black, and Ruslan Salakhutdinov. 2020.
\newblock \href {https://doi.org/10.18653/v1/2020.coling-main.1} {Exploring
  controllable text generation techniques}.
\newblock In \emph{Proceedings of the 28th International Conference on
  Computational Linguistics}, pages 1--14, Barcelona, Spain (Online).
  International Committee on Computational Linguistics.

\bibitem[{Radford et~al.(2019)Radford, Wu, Child, Luan, Amodei, Sutskever
  et~al.}]{radford2019language}
Alec Radford, Jeffrey Wu, Rewon Child, David Luan, Dario Amodei, Ilya
  Sutskever, et~al. 2019.
\newblock Language models are unsupervised multitask learners.
\newblock \emph{OpenAI blog}, 1(8):9.

\bibitem[{Raffel et~al.(2020)Raffel, Shazeer, Roberts, Lee, Narang, Matena,
  Zhou, Li, and Liu}]{raffel2020exploring}
Colin Raffel, Noam Shazeer, Adam Roberts, Katherine Lee, Sharan Narang, Michael
  Matena, Yanqi Zhou, Wei Li, and Peter~J Liu. 2020.
\newblock Exploring the limits of transfer learning with a unified text-to-text
  transformer.
\newblock \emph{Journal of Machine Learning Research}, 21:1--67.

\bibitem[{Rao and Tetreault(2018)}]{rao-tetreault-2018-dear}
Sudha Rao and Joel Tetreault. 2018.
\newblock \href {https://doi.org/10.18653/v1/N18-1012} {Dear sir or madam, may
  {I} introduce the {GYAFC} dataset: Corpus, benchmarks and metrics for
  formality style transfer}.
\newblock In \emph{Proceedings of the 2018 Conference of the North {A}merican
  Chapter of the Association for Computational Linguistics: Human Language
  Technologies, Volume 1 (Long Papers)}, pages 129--140, New Orleans,
  Louisiana. Association for Computational Linguistics.

\bibitem[{Sohn et~al.(2015)Sohn, Lee, and Yan}]{DBLP:conf/nips/SohnLY15}
Kihyuk Sohn, Honglak Lee, and Xinchen Yan. 2015.
\newblock \href
  {https://proceedings.neurips.cc/paper/2015/hash/8d55a249e6baa5c06772297520da2051-Abstract.html}
  {Learning structured output representation using deep conditional generative
  models}.
\newblock In \emph{Advances in Neural Information Processing Systems 28: Annual
  Conference on Neural Information Processing Systems 2015, December 7-12,
  2015, Montreal, Quebec, Canada}, pages 3483--3491.

\bibitem[{Sun et~al.(2022)Sun, Chen, Zhou, Li, Cao, and Zheng}]{sun2022non}
Rongyi Sun, Borun Chen, Qingyu Zhou, Yinghui Li, Yunbo Cao, and Hai-Tao Zheng.
  2022.
\newblock A non-hierarchical attention network with modality dropout for
  textual response generation in multimodal dialogue systems.
\newblock In \emph{ICASSP 2022-2022 IEEE International Conference on Acoustics,
  Speech and Signal Processing (ICASSP)}, pages 6582--6586. IEEE.

\bibitem[{Vijayakumar et~al.(2016)Vijayakumar, Cogswell, Selvaraju, Sun, Lee,
  Crandall, and Batra}]{DBLP:journals/corr/VijayakumarCSSL16}
Ashwin~K. Vijayakumar, Michael Cogswell, Ramprasaath~R. Selvaraju, Qing Sun,
  Stefan Lee, David~J. Crandall, and Dhruv Batra. 2016.
\newblock \href {http://arxiv.org/abs/1610.02424} {Diverse beam search:
  Decoding diverse solutions from neural sequence models}.
\newblock \emph{CoRR}, abs/1610.02424.

\bibitem[{Wang et~al.(2019)Wang, Wu, Mou, Li, and Chao}]{wang2019harnessing}
Yunli Wang, Yu~Wu, Lili Mou, Zhoujun Li, and Wenhan Chao. 2019.
\newblock Harnessing pre-trained neural networks with rules for formality style
  transfer.
\newblock In \emph{Proceedings of the 2019 Conference on Empirical Methods in
  Natural Language Processing and the 9th International Joint Conference on
  Natural Language Processing (EMNLP-IJCNLP)}, pages 3573--3578.

\bibitem[{Xu et~al.(2020)Xu, Patwary, Shoeybi, Puri, Fung, Anandkumar, and
  Catanzaro}]{xu2020megatron}
Peng Xu, Mostofa Patwary, Mohammad Shoeybi, Raul Puri, Pascale Fung, Animashree
  Anandkumar, and Bryan Catanzaro. 2020.
\newblock Megatron-cntrl: Controllable story generation with external knowledge
  using large-scale language models.
\newblock In \emph{Proceedings of the 2020 Conference on Empirical Methods in
  Natural Language Processing (EMNLP)}, pages 2831--2845.

\bibitem[{Xu et~al.(2019)Xu, He, Su, Zhong, Xu, Gu, and Huang}]{8946478}
Shilin Xu, Zhimin He, Junjian Su, Liangsheng Zhong, Yue Xu, Huimin Gu, and
  Yubing Huang. 2019.
\newblock \href {https://doi.org/10.1109/ICWAPR48189.2019.8946478} {A shopping
  guide text generation system based on deep neural network}.
\newblock In \emph{2019 International Conference on Wavelet Analysis and
  Pattern Recognition (ICWAPR)}, pages 1--5.

\bibitem[{Yi et~al.(2020)Yi, Li, Yang, Li, and Sun}]{yi2020mixpoet}
Xiaoyuan Yi, Ruoyu Li, Cheng Yang, Wenhao Li, and Maosong Sun. 2020.
\newblock Mixpoet: Diverse poetry generation via learning controllable mixed
  latent space.
\newblock In \emph{Proceedings of the AAAI conference on artificial
  intelligence}, volume~34, pages 9450--9457.

\bibitem[{Zhang et~al.(2018)Zhang, Dinan, Urbanek, Szlam, Kiela, and
  Weston}]{zhang2018personalizing}
Saizheng Zhang, Emily Dinan, Jack Urbanek, Arthur Szlam, Douwe Kiela, and Jason
  Weston. 2018.
\newblock Personalizing dialogue agents: I have a dog, do you have pets too?
\newblock In \emph{Proceedings of the 56th Annual Meeting of the Association
  for Computational Linguistics (Volume 1: Long Papers)}, pages 2204--2213.

\bibitem[{Zheng et~al.(2020)Zheng, Cai, Zhang, and Li}]{zheng2020controllable}
Changmeng Zheng, Yi~Cai, Guanjie Zhang, and Qing Li. 2020.
\newblock Controllable abstractive sentence summarization with guiding
  entities.
\newblock In \emph{Proceedings of the 28th International Conference on
  Computational Linguistics}, pages 5668--5678.

\end{thebibliography}
\bibliographystyle{acl_natbib}

\clearpage
\appendix

\section{Appendix}
\label{sec:appendix}

\subsection{Dataset Details}\label{sec:datadetail}

The size of each object/occasion category is shown in Table~\ref{tab:obj} and Table~\ref{tab:occ} respectively. 
It is worth noting that the “General” category refers to the case where the sending object of corresponding blessing is not acquired during the data collection process.
In addition, there is mutual inclusion between some objects in our dataset. We consider this phenomenon is reasonable, e.g., we may write only one blessing message for elders, and send it to others, such as parents, uncles, and teachers, with a little modification.

Some examples of \DatasetName{} are shown in Table~\ref{table:samples} which contain the blessings and the corresponding attributes (i.e., occasions and objects).

\begin{table}[ht]
\centering
\small
\begin{tabular}{cc|cc}
\toprule
Object     & Size   & Object        & Size \\
\midrule
General    & 102,284 & Customer      & 4,220 \\
Friend     & 48,058  & Parent        & 3,791 \\
Lover      & 20,314  & Newlywed      & 3,210 \\
Teacher    & 14,092  & Senior        & 2,785 \\
Girlfriend & 9,323   & Daughter      & 2,614 \\
Dad        & 8,871   & Son           & 2,610 \\
Kid        & 8,373   & Employee      & 2,272 \\
Wife       & 7,584   & Grandma       & 427  \\
Husband    & 7,478   & Cousin        & 403  \\
Boyfriend  & 6,603   & Niece         & 319  \\
Student    & 6,600   & Granddaughter & 238  \\
Boss       & 5,592   & Aunt          & 215  \\
Sister     & 5,208   & Grandson      & 210  \\
Colleague  & 4,896   & Nephew        & 160  \\
Brother    & 4,884   & Grandpa       & 155  \\
Classmate  & 4,770   & Uncle         & 135  \\
Mom        & 4,602   & Grandparent   & 107  \\
\bottomrule
\end{tabular}
\caption{The data size of each object category.}
\label{tab:obj}
\end{table}

\begin{table}[ht]
\centering
\small
\begin{tabular}{cc|cc}
\toprule
Occasion        & Size  & Occasion         & Size \\
\midrule
New Year      & 55,162 & Farewell       & 9,046 \\
Birthday      & 36,329 & Valentine's Day & 8,727 \\
Christmas     & 27,713 & Halloween      & 6,050 \\
Wedding       & 21,039 & Mother's Day    & 4,132 \\
Good Morning  & 18,197 & Exam           & 3,659 \\
Thanksgiving  & 15,245 & Happy Weekend  & 3,506 \\
Graduation    & 15,234 & Good Afternoon & 2,284 \\
Father's Day   & 14,344 & Fool's Day      & 2,074 \\
Teacher's Day  & 12,321 & Easter         & 1,999 \\
Good Night    & 11,667 & Housewarming   & 1,810 \\
Children's Day & 11,067 & Women's Day     & 1,239 \\
Anniversary   & 10,559 &                &     \\
\bottomrule
\end{tabular}
\caption{The data size of each occasion category.}
\label{tab:occ}
\end{table}

\begin{table*}[ht]
\centering
\small
\begin{tabular}  {p{2\columnwidth}}
\toprule
\textbf{\textcolor{red}{[Anniversary]} \textcolor{blue}{[Aunt]}} \textcolor{red}{Happy Anniversary} to the people \textcolor{blue}{I look up to} whenever I am in doubt. Dear \textcolor{blue}{uncle and aunty}, you guys are surely made for each other. \textcolor{red}{Have a great year ahead}.
\\\midrule
\textbf{\textcolor{red}{[Anniversary]} 
\textcolor{blue}{[Parents]}} You are the \textcolor{blue}{parents} that all \textcolor{blue}{kids} hope to have, you are the \textcolor{red}{couple} that all \textcolor{red}{lovers} hope to be and you both are the \textcolor{blue}{pillars of support} that every \textcolor{blue}{family} wishes it had. \textcolor{red}{Happy anniversary} to the best \textcolor{blue}{parents} ever.
\\\midrule
\textbf{\textcolor{red}{[Birthday]} 
\textcolor{blue}{[Daughter]}}
This day is truly \textcolor{red}{a special day} for us because this is the day when we \textcolor{red}{first had a glimpse} on \textcolor{blue}{our angel}. \textcolor{red}{Have a lovely birthday} our \textcolor{blue}{dear daughter}!
\\\midrule
\textbf{\textcolor{red}{[Birthday]} 
\textcolor{blue}{[Colleague]}}
\textcolor{red}{Ignite one candle}, happiness will last forever. We \textcolor{blue}{work together} for around one-third in a day, and it’s more than we spend time with our family, and It means \textcolor{blue}{we are colleagues}. \textcolor{red}{Happy birthday}, \textcolor{blue}{dear colleague}!
\\\midrule
\textbf{\textcolor{red}{[Children's Day]} 
\textcolor{blue}{[Kid]}}
\textcolor{blue}{My child}, I bless you on this special day. You will never \textcolor{blue}{grow up}, I wish you \textcolor{red}{a happy Children's Day}!
\\\midrule
\textbf{\textcolor{red}{[Children's Day]} 
\textcolor{blue}{[Student]}}
We may be your \textcolor{blue}{teachers} but we also have a lot more things to \textcolor{blue}{learn from} you, especially how to \textcolor{red}{laugh with all your hearts}. \textcolor{red}{Happy children’s day}!
\\\midrule
\textbf{\textcolor{red}{[Christmas]} \textcolor{blue}{[Boss]}} \textcolor{red}{Merry Christmas} to a \textcolor{blue}{boss} who keeps the \textcolor{blue}{office} \textcolor{red}{humming} along like \textcolor{red}{Santa}’s \textcolor{blue}{Workshop}! \\\midrule
\textbf{\textcolor{red}{[Christmas]} \textcolor{blue}{[Wife]}} \textcolor{blue}{Precious wife}, my \textcolor{blue}{heart} hangs on your every breath, like lights hanging on a \textcolor{red}{Christmas tree}. \textcolor{red}{Merry Christmas} my \textcolor{blue}{dear love}! \\\midrule

\textbf{\textcolor{red}{[Easter]} \textcolor{blue}{[Boyfriend]}}
One the beautiful \textcolor{red}{Easter day}, \textcolor{blue}{my boyfriend}, let \textcolor{red}{the prayers and fasting for Lord Jesus} bring much \textcolor{blue}{love and happiness} in our lives. I \textcolor{red}{pray to the Lord} to make \textcolor{blue}{our relationship} \textcolor{red}{fruitful and prosperous}. \textcolor{red}{Have a happy Easter}.
\\\midrule
\textbf{\textcolor{red}{[Easter]} \textcolor{blue}{[Teacher]}}
\textcolor{blue}{Dear teacher}, you are \textcolor{blue}{my inspiration} and I am happy to be \textcolor{blue}{under your guidance}. It’s such \textcolor{red}{a hopeful time of year}, I hope your heart gets filled with love, \textcolor{red}{baskets with candies} and \textcolor{red}{Easter eggs this Easter}.
\\\midrule
\textbf{\textcolor{red}{[Thanksgiving]} \textcolor{blue}{[Boss]}} It has always been a pleasure \textcolor{blue}{working} with you because it has been great \textcolor{blue}{learning}. \textcolor{red}{Thanking you} for playing a \textcolor{blue}{leading} role in my happiness at \textcolor{blue}{work}. Warm greetings on \textcolor{red}{Thanksgiving}!
\\\midrule
\textbf{\textcolor{red}{[Thanksgiving]} \textcolor{blue}{[Wife]}} Today, I want to \textcolor{red}{say thanks} so much for accepting to spend the rest of your life with me. \textcolor{red}{Thanks} so much for being my \textcolor{blue}{heart beat}. I \textcolor{blue}{love} you, \textcolor{blue}{dear wife}. Happy \textcolor{red}{Thanksgiving day}!\\
\bottomrule
\end{tabular}\caption{The examples of \DatasetName{}. The words related to the attribute \textbf{\textcolor{red}{Occasion}} and \textbf{\textcolor{blue}{Object}} are highlighted(e.g. \textcolor{red}{Happy Anniversary}).}
\label{table:samples}
\end{table*}

\begin{table*}[ht]
\small
\centering
\begin{tabular}{lccc|lccc}
\toprule
Object & \#Sample & \#Per. & \#Ent. & Occasion & \#Sample & \#Per. & \#Ent. \\ \midrule
Aunt & 80 & 73 & 66 & Anniversary & 260 & 239 & 214 \\
Boss & 140 & 126 & 118 & Birthday & 520 & 479 & 436 \\
Boyfriend & 240 & 223 & 197 & Children's Day & 120 & 112 & 104 \\
Brother & 220 & 199 & 180 & Christmas & 480 & 445 & 397 \\
Classmate & 220 & 206 & 183 & Easter & 100 & 92 & 81 \\
Colleague & 200 & 187 & 162 & Exam & 200 & 185 & 163 \\
Cousin & 60 & 55 & 48 & Farewell & 220 & 205 & 186 \\
Customer & 100 & 90 & 83 & Father's Day & 120 & 112 & 99 \\
Dad & 140 & 128 & 115 & Fool's Day & 40 & 38 & 35 \\
Daughter & 200 & 188 & 168 & Good Afternoon & 120 & 113 & 100 \\
Employee & 120 & 112 & 97 & Good Morning & 340 & 310 & 279 \\
Friend & 440 & 405 & 361 & Good Night & 260 & 239 & 213 \\
Girlfriend & 300 & 278 & 248 & Graduation & 300 & 285 & 243 \\
Granddaughter & 60 & 55 & 52 & Halloween & 160 & 144 & 134 \\
Grandma & 40 & 38 & 33 & Happy Weekend & 60 & 56 & 48 \\
Grandpa & 20 & 18 & 16 & Housewarming & 60 & 53 & 50 \\
Grandparent & 20 & 17 & 14 & Mother's Day & 140 & 131 & 114 \\
Grandson & 40 & 37 & 31 & New Year & 440 & 404 & 368 \\
Husband & 260 & 237 & 211 & Teacher's Day & 80 & 76 & 63 \\
Kid & 180 & 168 & 145 & Thanksgiving & 320 & 300 & 260 \\
Lover & 340 & 319 & 282 & Valentine's Day & 300 & 274 & 245 \\
Mom & 200 & 188 & 165 & Wedding & 320 & 289 & 262 \\
Nephew & 20 & 19 & 17 & Women's Day & 100 & 95 & 76 \\
Newlywed & 20 & 18 & 16 &  &  &  &  \\
Niece & 60 & 56 & 51 &  &  &  &  \\
Parent & 160 & 147 & 133 &  &  &  &  \\
Senior & 80 & 73 & 69 &  &  &  &  \\
Sister & 240 & 216 & 195 &  &  &  &  \\
Son & 200 & 189 & 166 &  &  &  &  \\
Student & 160 & 149 & 130 &  &  &  &  \\
Teacher & 200 & 181 & 164 &  &  &  &  \\
Uncle & 40 & 37 & 33 &  &  &  &  \\
Wife & 260 & 244 & 221 &  &  &  & \\
\bottomrule
\end{tabular}
\caption{Complete human evaluation results of~\DatasetName{}. \#Sample, \#Per. and \#Ent. denote the total number of sampled sentences, the number of personalized sentences and the number of entangled sentences respectively.}
\label{tab:human2}
\end{table*}

\begin{table*}[ht]
\centering
\small
\begin{tabular} {lp{1.65\columnwidth}}
\toprule
Object/Occasion & Related Words \\ \midrule
Colleague & colleague, work, workplace, office, workshop, companion, workmate, coworker, mate, associate, helper, partner, hard, company, career, wealth, business \\ \midrule
Boss & boss, work, workplace, office, workshop, chairman, chief, head, sir, supervisor, leader, charge, administrator, management, leadership, dictator, rule, thank, success, company, full, career, team, support, help, guidance, money, business, mentor, job, employees, create, development, professional, encouragement, achievements \\ \midrule
Girlfriend & girlfriend, queen, love, addicted, kiss, sweetheart, mate, bestie, date, babe, baby, partner, forever, heart, beautiful, dear, sweet, forever, together, give, long, dreams, warm, sun, care, thank, future, moment, wind, bright, gift, remember, lovely, honey, promise, cherish, promise, shining, flower \\ \midrule
Aunt & aunt, uncle, aunty, dear, sweet, family \\ \midrule
Boyfriend & boyfriend, love, addicted, kiss, sweetheart, mate, bestie, date, babe, baby, partner, heart, forever, darling, honey, promise, charming, precious, hugs, accompany \\ \midrule
Brother & brother, dear, forever, sister, sweet, heart, luck, joy, proud, engagement, health, family, harmony, thanks, handsome, follow, childhood, room \\ \midrule
Classmate & classmate, friend, together, forever, long, sincere, graduation, cherish, youth, road, think, school, everyone, memory, success, grow, accompany \\ \midrule
Cousin & cousin, grow, hand, family, forever \\ \midrule
Customer & new, customer, work, health, joy, friendship, money, client, gifts \\ \midrule
Dad & father, dad, love, thank, hard, warm, care, healthy, forever, family, dear, work, rain, give, young, strong, back, child, grow, son, daughter, support, parents, longevity, gratitude, kindness, umbrella, gentle, teaching, understand, journey, lamp, encouragement, illuminating, handsome, stalwart \\ \midrule
Daughter & daughter, dear, sweet, baby, family, princess, enjoy, gift, lovely, parents, born, th \\ \midrule
Employee & work, employee, thank, luck, future, success, career, dedication, together, forever, colleagues, team, success, appreciate, office \\ \midrule
Friend & happiness, friend, forever, dear, joy, friendship, warm, work, smile, care, sun, miss, sincerely, reunion, help, grow, accompany, kind, cherish, sunshine, gratitude, drink, successful, buddy, embrace, invite, lonely \\ \midrule
Granddaughter & granddaughter, dear, candies, sweet, grandpa, grandma, favorite, happy, trick, toy, beautiful \\ \midrule
Grandma & grandma, health, longevity, dear, old, thank, grandmother, joy, beautiful, sweet, kind \\ \midrule
Grandpa & grandpa, healthy, longevity, heart, smile, dear, old, thank, grandfather, joy, beautiful, sweet, kind, grandson, embrace \\ \midrule
Grandparents & grandparents, healthy, longevity, heart, smile, dear, old, thank, joy, beautiful, sweet, kind, grandson, embrace \\ \midrule
Grandson & grandson, cute, dear, candies, sweet, grandpa, grandma, favorite, happy, trick, toy, handsome, magic \\ \midrule
Husband & love, husband, dear, heart, life, always, thank, only, father, special, sweet, everything, marriage, wife, honey, baby, children, grateful, family, kind, wedding, marriage, cherish, met, deep, promise, moments, engagement \\ \midrule
Kid & children, kid, happy, childhood, childlike, innocence, child, little, heart, face, fun, growth, dreams, laugh, play, enjoy, haha, colorful, free, fly, lively \\ \midrule
Lover & love, heart, life, dear, sweet, forever, dreams, together, sweetheart, thank, wife, husband, honey, moment, light, warm, babe, cherish, promise, sure, met, shining, angels, partner, hug, breath, important \\ \midrule
Mom & mom, mother, love, thank, health, hard, forever, son, woman, daughter, grateful, parents, kindness \\ \midrule
Nephew & nephew, success, future, dear, life, proud, achieve, niece, adult \\ \midrule
Newlywed & newlywed, love, together, wedding, life, marriage, beautiful, new, congratulations, hundred, harmony, wife, pair, bridegroom, moment, phoenix, candles \\ \midrule
Niece & success, future, dear, life, proud, achieve, niece, adult, hard, beauty \\ \midrule
Parent & mom, parent, care, mother, family, father, life, thank, grateful, forever, children, warm, dear \\ \midrule
Senior & health, senior, old, long, longevity, thank, care, wealth, give, sir \\ \midrule
Sister & sister, dear, beautiful, heart, brother, old, little, gift, proud, family \\ \bottomrule
\end{tabular}
\caption{Bag-of-words related to objects and occasions.}
\label{tab:bag}
\end{table*}

\begin{table*}[ht]
\centering
\small
\begin{tabular} {lp{1.68\columnwidth}}
\toprule
Object/Occasion & Related Words \\ \midrule
Son & son, dear, sweet, baby, family, prince, enjoy, follow, handsome, gift, pride, lovely, parents, born, th \\ \midrule
Student & student, children, future, childhood, friends, innocence, childlike, smile, classmates, knowledge, grow, road, college, proud, career, study, wisdom, university, achieve, examination \\ \midrule
Teacher & teacher, hard, students, full, knowledge, care, light, podium, gratitude, chalk, soul, sun, wisdom, thank, kindness, forward, tree, dreams, education, support, learning, accompany, class, illuminating, wings, guidance, watering \\ \midrule
Uncle & uncle, old, aunt, dear, sweet, family \\ \midrule
Wife & wife, love, life, beautiful, heart, dear, mother, woman, everything, family, darling, warm, sweetheart, moment, thanks, dream, children, married, accompany, sunshine, given, deserve, help \\ \midrule
Christmas & Christmas, Xmas, merry, santa, card, tree, eve, stocking, humming, peace, claus, warm, night, gift, bell, snow, cold, deer, candlelight, chimney, elk, sled, shining, jesus \\ \midrule
Thanksgiving & thanksgiving, thank, grateful, gratitude, care, give, warm, smile, kindness, bright, cherish \\ \midrule
Graduation & graduation, congratulation, graduate, determination, dedication, achievement, life, future, success, proud, work, teacher, school, classmates, dreams, youth, journey, forward, college, leave, knowledge, continue, grow, study, society, examination \\ \midrule
Anniversary & wedding, love, increase, darling, marriage, th, year, couple, life, together, best, believe, more, wonderful, always, heart, wife, husband, long, future, relationship, sweet \\ \midrule
Birthday & birthday, happy, years, health, long, forever, special, gift, dreams \\ \midrule
Children's Day & children, childhood, always, sweet, play, innocent, smile, june, forever, face, young, grow \\ \midrule
Easter & easter, god, christ, lord, resurrection, new, eggs, spring, pray, basket, renewal, prosperity, bunny, rejoice, risen, holy \\ \midrule
Exam & exam, success, luck, god, pray, comes, write, believe, sure, result, grades, final, proud, study, lord, pass, wisdom, efforts, questions, victory, preparation, excellent, paper, deserve, confidence \\ \midrule
Farewell & farewell, life, goodbye, thank, friend, future, miss, luck, again, back, remember, memories, leaving, cherish, years \\ \midrule
Father's Day & father, dad, love, thank, mountain, sea, deep, strong, support, son, shoulders, strength, light, parents, tired, accompany, busy, gentle, umbrella, teachings, given, heavy \\ \midrule
Fool's Day & fool, april, happy, stupid, look, phone, money, really, smile, read, haha \\ \midrule
Good Afternoon & afternoon, day, enjoy, everything, sunshine, lunch, midday, relaxing, breath \\ \midrule
Good Morning & morning, face, new, smile, sun, start, light, mood, yesterday, embrace \\ \midrule
Good Night & night, sleep, goodnight, dreams, tomorrow, sweet, stars, pray, today, bed, moon, close, asleep, sound, amen \\ \midrule
Halloween & halloween, ghost, fun, pumpkin, afraid, lantern, candy, mask, witches, broom, moon, children, vampires, monster \\ \midrule
Happy Weekend & weekend, work, fun, relax, saturday, busy, enjoy, rest, tired, sleep \\ \midrule
Housewarming & new, house, housewarming, move, congratulations, come, neighbors, firecrackers, welcome \\ \midrule
Mother's Day & mother, love, thank, children, women, daughter, son, giving, grow, sea, raising, sunshine, breeze, embrace \\ \midrule
New Year & new, spring, coming, eve, change, warm, year, red, together, fireworks, welcome, forward, bright, prosperity, winter, busy, snow, cold, bloom, approaching, continue \\ \midrule
Teacher's Day & teacher, thank, work, students, knowledge, full, flowers, light, chalk, podium, sun, warm, candle, dedication, school, september, growth, tree, garden, respect, illuminate, education, children, classroom, guidance, ignited \\ \midrule
Valentine's Day & valentine, love, heart, dear, darling, together, sweet, promise, honey, share, romantic, handsome, beautiful, kiss, important, partner, babe \\ \midrule
Wedding & love, wedding, life, forever, marriage, congratulations, sweet, future, bride, harmony, wife, always, year, fate, home, share, moment \\ \midrule
Women's Day & women, beautiful, special, strength, wife, work, power, inspiration, proud, deserve, queen \\ \bottomrule
\end{tabular}
\caption{Bag-of-words related to objects and occasions.}
\label{tab:bag2}
\end{table*}

\end{document}